\documentclass[11pt]{article}

\usepackage[preprint]{acl}

\usepackage{times}
\usepackage{latexsym}

\usepackage[T1]{fontenc}

\usepackage{microtype}

\usepackage{inconsolata}

\usepackage{graphicx}

\usepackage{amsmath}
\usepackage{amssymb}
\usepackage{subcaption}
\usepackage{float}
\usepackage[most]{tcolorbox}
\usepackage{booktabs}
\usepackage{tabularx}
\usepackage{array}
\usepackage{xcolor}
\usepackage{colortbl}
\usepackage{multirow}
\usepackage{placeins}
\usepackage{fancyvrb}
\usepackage{algorithm}
\usepackage{algpseudocode}

\title{MVSS: A Unified Framework for Multi-View Structured Survey Generation}

\author{
  \textbf{Yinqi Liu}\textsuperscript{1},
  \textbf{Yueqi Zhu}\textsuperscript{1},
  \textbf{Yongkang Zhang}\textsuperscript{1},
  \textbf{Feiran Liu}\textsuperscript{1}, 
  \textbf{Yutong Shen}\textsuperscript{1}, \\
  \textbf{Yufei Sun}\textsuperscript{1},
  \textbf{Xin Wang}\textsuperscript{2},
  \textbf{Renzhao Liang}\textsuperscript{3}, 
  \textbf{Yidong Wang}\textsuperscript{2}, 
  \textbf{Cunxiang Wang}\textsuperscript{4}\thanks{~~Corresponding author.} \\
  \textsuperscript{1}Beijing University of Technology \quad
  \textsuperscript{2}Peking University \\
  \textsuperscript{3}Beihang University \quad
  \textsuperscript{4}Tsinghua University \\
  \texttt{liuyinqi726@gmail.com} \quad \texttt{wangcunxiang303@gmail.com}
}

\begin{document}
\maketitle

\begin{abstract}
Scientific surveys require not only summarizing large bodies of literature, but also organizing them into clear and coherent conceptual structures. 
However, existing automatic survey generation methods typically focus on linear text generation and struggle to explicitly model hierarchical relations among research topics and structured methodological comparisons, resulting in substantial gaps in structural organization and evidence presentation compared to expert-written surveys. To address this limitation, we propose \textbf{MVSS}, a \emph{multi-view structured survey generation} framework that jointly generates and aligns citation-grounded hierarchical trees, structured comparison tables, and survey text.
MVSS follows a structure-first paradigm: it first constructs a tree that captures the conceptual organization of a research domain, then generates comparison tables constrained by the tree structure, and finally uses both the tree and tables as joint structural constraints to guide outline construction and survey text generation.
This design enables complementary and aligned multi-view representations across structure, comparison, and narrative. In addition, we introduce a dedicated evaluation framework that systematically assesses generated surveys from multiple dimensions, including structural quality, comparative completeness, and citation fidelity.
Through large-scale experiments on \textbf{76} computer science topics, we demonstrate that MVSS significantly outperforms existing methods in survey organization and evidence grounding, and achieves performance comparable to expert-written surveys across multiple evaluation metrics.
\end{abstract}

\makeatletter
\renewcommand{\dbltopfraction}{0.95}
\renewcommand{\dblfloatpagefraction}{0.85}
\renewcommand{\textfraction}{0.05}
\renewcommand{\floatpagefraction}{0.85}
\makeatother
\section{Introduction}

\begin{figure*}[t]
    \centering
    \includegraphics[width=\textwidth, height=0.16\textheight,keepaspectratio]{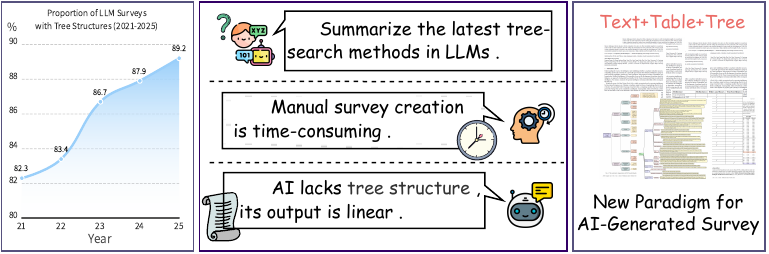}
    \caption{\textbf{Motivation for structure-first survey generation.}
    Linear-text survey generation obscures evolving comparison dimensions, motivating aligned multi-view structures.
    Expert surveys predominantly adopt explicit hierarchies (2021--2025), supporting a structure-first perspective.}
    \label{fig:motivation_overview}
\end{figure*}

In several frontier areas of natural language processing, methodological innovations are advancing at a pace that far outstrips the ability of manually written surveys to keep up.
In rapidly evolving domains such as large language models, retrieval-augmented generation, and multimodal reasoning~\citep{brown2020fewshot,openai2023gpt4,zhao2023surveyllm,lewis2020rag,gao2023ragsurvey,yin2024mllmsurvey}, major breakthroughs often emerge within months, while comprehensive surveys typically lag behind by one or more publication cycles.

Recently, LLM-based automated survey generation agents have attracted growing attention in both academia and industry.
Representative academic systems include AutoSurvey~\citep{huang2020autosurvey,wang2024autosurvey}, SurveyGen~\citep{bao2025surveygen}, SurveyX~\citep{liang2025surveyx}, SurveyForge~\citep{yan2025surveyforge}, SurveyG~\citep{nguyen2025surveyg}, InteractiveSurvey~\citep{wen2025interactivesurvey}, LLMxMapReduce~\citep{wang2025llmxmapreduce}, and related agent-based approaches~\citep{qi2025llmmas,ali2024autolitreview}.
In parallel, industrial research assistants such as OpenAI's Deep Research~\citep{openai2024deepresearch} and Google Gemini~\citep{google2023gemini} have demonstrated strong capabilities in large-scale literature understanding and synthesis.
However, the majority of these systems generate surveys primarily as linear narratives, overlooking other critical dimensions commonly present in human-written surveys—most notably citation trees and citation tables.

A \emph{citation tree} is a hierarchical structure used to organize survey content, where each node corresponds to a research concept or sub-area and is explicitly associated with a set of representative supporting papers.
The hierarchical relations between nodes reflect the progressive decomposition of a research topic from high-level themes to finer-grained directions.
A \emph{citation table}, in contrast, provides a structured tabular representation of survey content, where each row corresponds to a method or research line, each column captures a key attribute or comparison dimension, and explicit citations link table entries to concrete evidence in the literature.
By analyzing surveys published between 2021 and 2025 in the large language model domain, we find that over 80\% of surveys include explicit hierarchical trees.
Moreover, user studies indicate that more than 99\% of readers tend to first consult trees or tables, rather than raw text, when reading survey papers.

Among existing methods, HiReview~\citep{hu2024hireview} constructs citation trees using clustering-based taxonomies derived primarily from paper titles.
Such title-driven retrieval often misses papers whose relevance is not explicitly reflected in their titles and fails to leverage information beyond surface-level metadata, leading to unstable and noisy hierarchies.
Furthermore, HiReview evaluates tree quality only indirectly through downstream survey text quality, rather than performing principled assessments of the taxonomy itself.
In human-written surveys, citation trees and comparison tables are highly correlated and jointly used to organize content, yet HiReview does not generate or reason over tables at all.

To address these limitations, we propose \textbf{Multi-View Structured Survey (MVSS)}, a framework that formulates survey generation as a \emph{joint structural generation} problem rather than a purely linear text generation pipeline.
To construct rich and well-founded hierarchies, we introduce \textbf{Hierarchical Knowledge Trees (HKT)}, a principled representation for modeling the conceptual organization of a research domain.
HKT explicitly externalizes the latent conceptual hierarchy in surveys as a citation-grounded tree, enabling systematic modeling of the branching evolution of ideas, the formation of subfields, and relationships among methodological paradigms.
Building on HKT, we further propose a structure-aware generation mechanism that logically anchors and constrains table construction, producing structured comparison tables that remain consistent with the overall survey in terms of conceptual level, comparison dimensions, and evidence support.
Finally, MVSS uses both trees and tables as joint structural constraints to guide outline construction and generate high-quality survey text.
From a technical perspective, MVSS is not a simple assembly of modules, but a reformulation of survey generation as a cross-view structural alignment problem.

We systematically evaluate MVSS on 76 computer science topics, comparing against AutoSurvey, HiReview, SurveyX, SurveyForge, expert-written surveys, and other recent state-of-the-art baselines. To ensure robust and objective assessment, we employ a multi-dimensional evaluation framework across multiple LLM judges (GPT-4o, Gemini, and DeepSeek). Across all settings, MVSS consistently achieves the strongest overall performance. In particular, our method demonstrates a substantial lead in structural organization and relevance over automated baselines, and frequently approaches or even exceeds the quality of expert-written surveys. Additional citation-level evaluations further show that MVSS maintains reliable evidence alignment, with citation precision and recall both exceeding 75\%. Overall, these results indicate that MVSS effectively bridges the gap between automated survey generation and expert-written surveys in both structural organization and evidence grounding.

\begin{figure*}[t]
    \centering
    \includegraphics[
  width=\textwidth,
  height=0.35\textheight,
  keepaspectratio
]{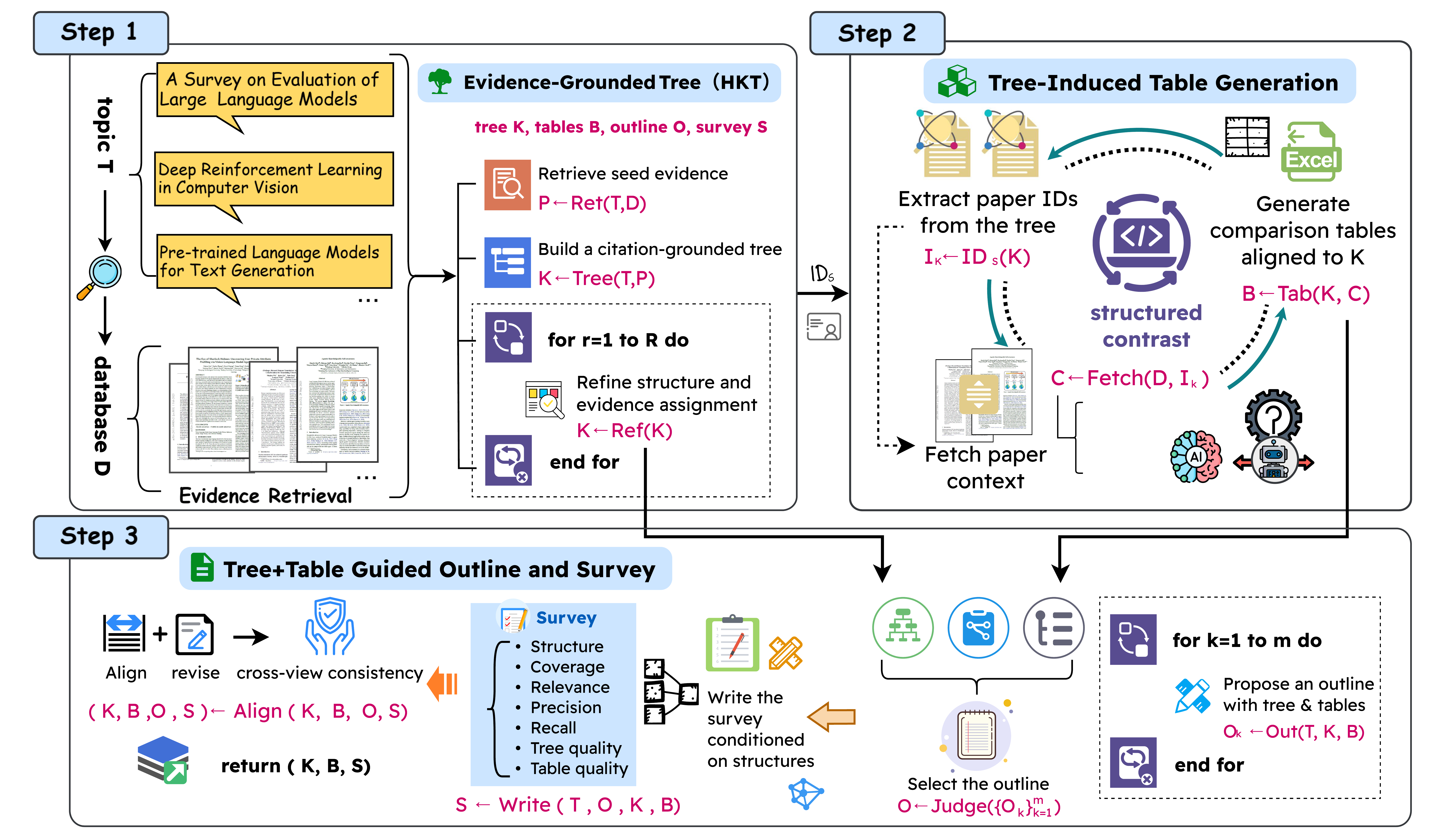}
    \caption{\textbf{Overview of \textsc{MVSS}.}
    Given a topic $T$ and a paper database $D$, \textsc{MVSS} constructs an evidence-grounded hierarchical tree, generates aligned comparison tables, and produces a structured survey via cross-view alignment.
}
    \label{fig:mvss_overview}
\end{figure*}

\section{Related Work}

\subsection{Automatic Survey Generation}

Automated survey generation increasingly assists literature reviews \citep{portenoy2020autoreview,kasanishi2023scireviewgen,darrin2024glimpse,bonorino2023smartsurveys}, mostly formulated as multi-document summarization producing linear narratives \citep{christensen2014summa,celikyilmaz2010hybridmds,liu2019hiersumm,yasunaga2019scisummnet,li2023hiermds,zhang2024tomds}. While some methods add structural biases \citep{liu2021highlighttransformer} or retrieval augmentation \citep{izacard2021fid,nogueira2019bertranker}, recent LLM systems emphasize citation-aware generation and critique-based refinement \citep{kryscinski2020factcc,dixit2023efactsum,kasanishi2023scireviewgen,madaan2023selfrefine,shinn2023reflexion}.

Despite these advances, existing systems remain fundamentally \emph{text-centric}, relying on intermediate artifacts unoptimized as explicit knowledge structures. MVSS instead adopts a \emph{structure-first} perspective, treating hierarchical trees, comparison tables, and cross-view alignment as primary optimization objectives rather than byproducts of text generation.

\begin{algorithm*}[t]
\caption{\textsc{MVSS}: Multi-View Structured Survey Generation}
\label{alg:mvss_3phase}
\begin{algorithmic}[1]
\Require topic $T$, paper database $D$
\Ensure tree $K$, tables $B$, outline $O$, survey $S$

\Statex \textbf{Phase 1: Evidence-Grounded Tree (HKT)}
\State Retrieve seed evidence: $P \gets \mathrm{Ret}(T, D)$
\State Build a citation-grounded tree: $K \gets \mathrm{Tree}(T, P)$
\For{$r=1$ to $R$}
    \State Refine structure and evidence assignment: $K \gets \mathrm{Ref}(K)$
\EndFor

\Statex \textbf{Phase 2: Tree-Induced Table Generation}
\State Extract paper IDs from the tree: $I_K \gets \mathrm{IDs}(K)$
\State Fetch paper context: $C \gets \mathrm{Fetch}(D, I_K)$
\State Generate comparison tables aligned to $K$: $B \gets \mathrm{Tab}(K, C)$

\Statex \textbf{Phase 3: Tree+Table Guided Outline and Survey}
\For{$k=1$ to $m$}
    \State Propose an outline with tree \& tables: $O_k \gets \mathrm{Out}(T, K, B)$
\EndFor
\State Select the outline: $O \gets \mathrm{Judge}(\{O_k\}_{k=1}^m)$
\State Write the survey conditioned on structures: $S \gets \mathrm{Write}(T, O, K, B)$
\State Align and revise for cross-view consistency: $(K,B,O,S) \gets \mathrm{Align}(K,B,O,S)$

\State \Return $(K,B,S)$
\end{algorithmic}
\end{algorithm*}

\subsection{Knowledge Structuring and Multi-Signal Evaluation}

Prior scientific knowledge organization relied on document embeddings \citep{beltagy2019scibert,cohan2020specter}, graph retrieval \citep{kasela2025kgretrieval}, and bibliometric clustering \citep{waltman2012classification}, yielding coarse structures insufficient for fine-grained surveys. Unlike recent LLM-based taxonomies that serve primarily as standalone exploration tools \citep{hsu2024chime,kargupta2025taxoadapt,zhu2025contextaware}, our Hierarchical Knowledge Tree (HKT) acts as a strict structural anchor that dynamically constrains table generation and narrative synthesis to ensure cross-view consistency.

Separately, multi-signal evaluation improves LLM reliability via reflection \citep{madaan2023selfrefine,shinn2023reflexion}, LLM-as-a-judge \citep{fu2024gptscore,bhat2023llmasannotators,bansal2023llmasannotators}, fact verification \citep{kryscinski2020factcc,dixit2023efactsum}, and reranking \citep{nogueira2019bertranker}. Yet, structure remains secondary. \textbf{MVSS bridges this gap by making structure the optimization target.} By jointly modeling hierarchical abstraction and verification, MVSS ensures trees, tables, and text form a coherent, faithfully grounded representation of the research landscape.
\section{Method}
\label{sec:method}

In this section, we describe the methodology of \textsc{MVSS}, a multi-view structured framework for automated survey generation.
MVSS formulates survey synthesis as a joint structural generation problem, where multiple representations are constructed and aligned in a coordinated manner.
Specifically, MVSS proceeds through three structured stages:
(1) evidence-grounded hierarchical knowledge tree construction,
(2) tree-induced structured table generation, and
(3) tree- and table-guided outline and survey text generation.
Each stage is designed to address a key challenge in automated survey creation, including conceptual organization, comparative analysis, and evidence-consistent writing.
Figure~\ref{fig:mvss_overview} provides an overview of the complete workflow, and the overall procedure is summarized in Algorithm~\ref{alg:mvss_3phase}.

\subsection{Evidence-Grounded Tree (HKT)}

Given a survey topic $T$ and a paper database $D$, MVSS first constructs a Hierarchical Knowledge Tree (HKT) to explicitly model the conceptual organization of the target domain.
We begin by retrieving an initial set of candidate papers:
\[
P \leftarrow \mathrm{Ret}(T, D),
\]
which serves as the evidence pool for subsequent structure induction.

Using the topic $T$ and retrieved papers $P$, we construct an initial tree:
\[
K \leftarrow \mathrm{Tree}(T, P),
\]
where each node corresponds to a research concept or subtopic and is explicitly associated with a set of supporting papers.
The resulting tree provides a structured abstraction of the domain, capturing major branches and their conceptual relations.

To improve structural stability and evidence consistency, we further apply an iterative refinement procedure:
\[
K \leftarrow \mathrm{Ref}(K),
\]
which adjusts the hierarchy by resolving redundant nodes, correcting parent--child relations, and reassigning supporting papers when necessary.
After $R$ refinement rounds, the tree $K$ serves as a citation-grounded conceptual backbone for subsequent stages.

\subsection{Tree-Induced Table Generation}

Based on the refined hierarchical knowledge tree $K$, MVSS generates structured comparison tables to explicitly expose discriminative dimensions among methods and subfields.
We first extract all paper identifiers associated with tree nodes:
\[
I_K \leftarrow \mathrm{IDs}(K),
\]
and retrieve their corresponding contextual information from the database:
\[
C \leftarrow \mathrm{Fetch}(D, I_K),
\]
including titles, abstracts, and other descriptive metadata.

Table generation is then formulated as a conditional mapping:
\[
B = \mathrm{Tab}(K, C),
\]
where the tree structure $K$ determines the semantic scope and placement of each table, and the paper context $C$ provides evidence for populating table entries.
Each table is aligned with a specific node or subtree, ensuring that comparison rows and columns operate at a consistent conceptual level.
As a result, the generated tables present structured, evidence-grounded comparisons that are coherent with the global survey organization.

\subsection{Tree+Table Guided Outline and Survey}

In the final stage, MVSS jointly leverages the hierarchical tree $K$ and comparison tables $B$ to guide outline construction and survey writing.
Given the topic and structural representations, we generate multiple outline candidates:
\[
O_k \leftarrow \mathrm{Out}(T, K, B), \quad k=1,\dots,m,
\]
and select them using a judge model:
\[
O \leftarrow \mathrm{Judge}(\{O_k\}_{k=1}^m).
\]

Conditioned on the final outline and structural constraints, the survey text is generated as:
\[
S = \mathrm{Write}(T, O, K, B),
\]
where the outline controls section ordering, the tree enforces conceptual hierarchy, and the tables guide comparative analysis and evidence usage.

To ensure consistency across multiple representations, we apply a cross-view alignment operation:
\[
(K,B,O,S) \leftarrow \mathrm{Align}(K,B,O,S),
\]
which detects and revises structural conflicts, missing coverage, or inconsistent citations among the tree, tables, outline, and text.
This alignment step yields a coherent, multi-view survey with consistent structure and evidence grounding.

\section{Experiments}

\subsection{Experimental Setup}

We conduct comprehensive experiments to evaluate MVSS and its hierarchical knowledge tree (HKT) module from three perspectives:
(1) quality of the generated knowledge trees,
(2) quality of the structured comparison tables, and
(3) overall survey quality when trees and tables are used jointly to guide text generation.

\paragraph{Dataset and Corpus.}
We evaluate on 76 computer science topics spanning machine learning, natural language processing, computer vision, and systems research.
The retrieval corpus contains 530{,}000 arXiv papers (2018--2024), preprocessed following standard practices for scientific document retrieval.
We use \textbf{deepseek-chat} as the primary generator to produce MVSS outputs.

\paragraph{LLM-as-judge evaluation.}
Following recent work on automatic evaluation of long-form generation, we rely on calibrated LLM judges to score trees, tables, and surveys.
All prompts are aligned with human-written guidelines, and a small expert-annotated set is used for scale calibration, echoing findings that large language models can serve as reliable automatic evaluators when appropriately designed and validated against human judgments~\citep{kocmi2023largelms,fabbri2021summeval}.

\subsection{Metrics}

We develop a new evaluation scheme tailored to multi-view structured survey generation.
It includes three dimensions for surveys (\textbf{Coverage}, \textbf{Structure}, \textbf{Relevance}), together with a single holistic criterion for trees (\textbf{TreeQuality}) and for tables (\textbf{TableQuality}).
All criteria are rated on a 1--5 Likert scale by calibrated LLM judges.

\paragraph{Survey quality.}
We evaluate surveys along three 5-point dimensions summarized in Table~\ref{tab:mvss_metrics}.
Coverage assesses whether the survey covers key and peripheral aspects comprehensively.
Structure evaluates logical organization, coherence, and non-redundant flow.
Relevance measures alignment with the target topic and focus with minimal digressions.
We define the overall survey score as $Q_{\text{survey}}=\frac{1}{3}\!\left(S_{\text{cov}}+S_{\text{str}}+S_{\text{rel}}\right)$.

\paragraph{Tree quality.}
TreeQuality evaluates the taxonomy/topic tree quality, including hierarchy correctness, coverage of major branches, and clarity of grouping.
We set $Q_{\text{tree}}=S_{\text{tq}}$.

\paragraph{Table quality.}
TableQuality evaluates the usefulness of comparison tables, focusing on correctness, completeness, consistency, and practical utility for comparison.
We set $Q_{\text{table}}=S_{\text{tab}}$.

\begin{table}[t]
\centering
\caption{New evaluation criteria for MVSS. All dimensions use a 1--5 Likert scale. For brevity, we show representative anchors (1/5).}
\label{tab:mvss_metrics}
\small
\renewcommand{\arraystretch}{1.00}
\begin{tabularx}{\columnwidth}{l>{\scriptsize}X}
\toprule
\textbf{Criterion} & \textbf{Anchors of 1--5 scale} \\
\midrule
\textbf{Coverage} &
1: Very limited coverage; misses most key areas.
5: Fully comprehensive; covers key and peripheral topics in depth. \\
\midrule
\textbf{Structure} &
1: No clear logic or connections between sections.
5: Tightly structured, clear logic, smooth transitions, no redundancy. \\
\midrule
\textbf{Relevance} &
1: Outdated/unrelated; not aligned with the topic.
5: Exceptionally focused; every detail supports understanding of the topic. \\
\midrule
\textbf{TreeQuality} &
1: No meaningful tree or totally wrong hierarchy.
5: Excellent tree: comprehensive, correct, clear grouping, useful abstraction. \\
\midrule
\textbf{TableQuality} &
1: No usable table or incorrect/misleading.
5: Excellent tables: comprehensive comparisons with consistent formatting. \\
\bottomrule
\end{tabularx}
\end{table}

\begin{table*}[t]
\centering
\caption{Comprehensive performance comparison between MVSS and state-of-the-art baselines across multiple LLM judges. \textbf{C}: Coverage, \textbf{S}: Structure, \textbf{R}: Relevance, \textbf{Avg}: Overall (C+S+R)/3. Human writing serves as a reference upper bound.} 
\label{tab:performance_comparison}
\renewcommand{\arraystretch}{1.2} 
\resizebox{\textwidth}{!}{%
\begin{tabular}{l | cccc | cccc | cccc}
\toprule
\textbf{Judge Model} & \multicolumn{4}{c|}{\textbf{GPT-4o}} & \multicolumn{4}{c|}{\textbf{Gemini-2.5-pro}} & \multicolumn{4}{c}{\textbf{DeepSeek-chat}} \\
\textbf{Method} & \textbf{Cov} & \textbf{Str} & \textbf{Rel} & \textbf{Avg} & \textbf{Cov} & \textbf{Str} & \textbf{Rel} & \textbf{Avg} & \textbf{Cov} & \textbf{Str} & \textbf{Rel} & \textbf{Avg} \\
\midrule
LLMxMapReduce     & 4.27 & 4.43 & 4.00 & 4.23 & 4.35 & 4.29 & 4.73 & 4.46 & 4.08 & \underline{4.10} & 4.00 & 4.06 \\
SurveyForge       & 4.13 & 4.05 & \textbf{4.58} & 4.25 & 4.03 & 3.38 & 4.66 & 4.02 & 4.05 & 3.95 & 4.05 & 4.02 \\
SurveyG           & 4.45 & \underline{4.76} & 3.98 & 4.40 & 4.64 & \textbf{4.80} & \underline{4.97} & \underline{4.80} & 4.10 & 4.00 & 4.50 & 4.20 \\
SurveyX           & 4.07 & 3.95 & 4.10 & 4.04 & 3.80 & 3.55 & 4.52 & 3.96 & 4.02 & 3.98 & 4.00 & 4.00 \\
InteractiveSurvey & 4.07 & 4.36 & 4.02 & 4.15 & 2.93 & 2.91 & 3.56 & 3.13 & 3.95 & 3.90 & 4.03 & 3.96 \\
AutoSurvey        & \underline{4.60} & 4.60 & 4.46 & \underline{4.55} & \underline{4.66} & 4.33 & 4.86 & 4.62 & \textbf{4.13} & 4.06 & \underline{4.48} & \underline{4.22} \\
HiReview          & 3.67 & 3.00 & 4.00 & 3.56 & 3.94 & 3.00 & 4.00 & 3.65 & 3.79 & 3.77 & 4.00 & 3.85 \\
\midrule
\textit{Human writing (Ref.)} & \textit{4.50} & \textit{5.00} & \textit{4.76} & \textit{4.75} & \textit{4.90} & \textit{4.62} & \textit{5.00} & \textit{4.84} & \textit{4.66} & \textit{4.50} & \textit{5.00} & \textit{4.72} \\
\textbf{MVSS (Ours)} & \textbf{4.86} & \textbf{4.80} & \underline{4.47} & \textbf{4.71} & \textbf{4.98} & \underline{4.45} & \textbf{4.99} & \textbf{4.81} & \underline{4.11} & \textbf{4.39} & \textbf{4.99} & \textbf{4.50} \\
\bottomrule
\end{tabular}%
}
\end{table*}

\paragraph{Citation quality for trees and surveys.}
Following scientific fact verification, we extract a set of claims $C=\{c_i\}_{i=1}^{m}$ and model-proposed (claim, reference) pairs $P=\{(c_i,r_j)\}$.
An NLI model $V(c,r)\in\{0,1\}$ returns 1 if $r$ supports $c$.
Citation recall and precision are
\begin{equation}
\mathrm{Rec}_{\text{cite}} = \frac{|\{c\in C:\exists r,(c,r)\in P,V(c,r)=1\}|}{|C|},
\end{equation}
\begin{equation}
\mathrm{Prec}_{\text{cite}} = \frac{|\{(c,r)\in P:V(c,r)=1\}|}{|P|}.
\end{equation}

\subsection{Baselines}

We compare MVSS against representative human and automatic survey-generation baselines:

\begin{itemize}
    \item \textbf{Human Experts.} Expert-written surveys with manually curated hierarchies, used as an upper bound.
    \item \textbf{AutoSurvey.} An automatic survey generator that produces an outline and expands it into text with sentence-level citations.
    \item \textbf{HiReview.} A hierarchical review system that retrieves a fixed set of papers and generates taxonomy-guided survey text.
    \item \textbf{LLMxMapReduce.} A baseline approach that processes retrieved papers in a MapReduce fashion to summarize and synthesize the final survey text.
    \item \textbf{SurveyForge.} An automated framework designed to generate structured academic surveys from literature by extracting and organizing key information.
    \item \textbf{SurveyG.} A generation baseline that structures multi-document summaries into comprehensive surveys.
    \item \textbf{SurveyX.} A representative state-of-the-art system for automated literature review and survey generation.
    \item \textbf{InteractiveSurvey.} A system that incorporates iterative or interactive mechanisms to build the survey structure and content with human-in-the-loop or iterative feedback.
\end{itemize}

\subsection{Main Results}

We now evaluate MVSS as a full survey generator and compare its performance against a comprehensive set of baselines. The key findings from our end-to-end survey evaluation are summarized as follows:

\begin{itemize}
    \item \textbf{MVSS consistently achieves the highest overall quality among all automatic baselines across different LLM judges.}
    As shown in Table~\ref{tab:performance_comparison}, MVSS consistently ranks first among all automated methods across GPT-4o, Gemini, and DeepSeek evaluations. For instance, under GPT-4o, MVSS achieves an overall score of $4.71$, outperforming strong baselines like AutoSurvey ($4.55$) and LLMxMapReduce ($4.23$). Similar advantages are observed with Gemini (scoring $4.81$), highlighting its model-agnostic superiority. Furthermore, paired $t$-tests confirm these improvements are statistically significant ($p < 0.05$, see Table~\ref{tab:statistical_significance}).

    \item \textbf{MVSS highly competitive with expert-level human writing.}
    Notably, MVSS is highly competitive with human-written surveys. Under GPT-4o and Gemini, MVSS's overall scores ($4.71$ and $4.81$) nearly match those of Human Experts ($4.75$ and $4.84$). Although DeepSeek rates human writing higher ($4.72$ vs. $4.50$), MVSS remains the closest automatic system to this upper bound, demonstrating its ability to synthesize literature with expert-level comprehensiveness and logical flow.

    \item \textbf{Structural organization is the primary source of MVSS's advantage.}
    Across all evaluations, MVSS maintains a substantial lead in the \textit{Structure} dimension over non-hierarchical baselines. For example, under GPT-4o, MVSS attains a Structure score of $4.80$, significantly outperforming LLMxMapReduce ($4.43$) and SurveyForge ($4.05$). This proves that explicit hierarchical modeling (via HKT) and cross-view alignment effectively eliminate redundancy and ensure coherent narrative flow, addressing the disorganized pitfalls of standard RAG models in long-form generation.
\end{itemize}

\subsection{Human Evaluation}
\label{sec:human_eval}

To complement our automated metrics, we conducted a double-blind human study across 76 topics. Expert annotators rated generated surveys on a 1--5 Likert scale across multiple dimensions to ensure qualitative rigor. As shown in Table~\ref{tab:human_evaluation}, MVSS significantly outperforms all automated baselines, achieving a superior overall score of $4.58 \pm 0.38$, which closely nears the human expert upper bound of $4.88$. This strong correlation between expert ratings and LLM-based judgments confirms that MVSS effectively produces structured, human-aligned surveys while validating the statistical reliability of our evaluation framework.

\begin{table}[htbp]
\centering
\caption{Human evaluation results comparing MVSS against state-of-the-art baselines and human experts. All dimensions are rated by humans on a 1--5 scale.}
\label{tab:human_evaluation}
\renewcommand{\arraystretch}{1.3} 
\setlength{\tabcolsep}{3.5pt}    
\resizebox{1.0\columnwidth}{!}{
\begin{tabular}{lcccc} 
\toprule
\textbf{Model} & \textbf{Coverage} & \textbf{Structure} & \textbf{Relevance} & \textbf{Overall Score} \\
\midrule
Human Expert & $4.87 \pm 0.29$ & $4.89 \pm 0.18$ & $4.88 \pm 0.20$ & $4.88 \pm 0.15$ \\
\textbf{MVSS (Ours)} & $\mathbf{4.62 \pm 0.45}$ & $\mathbf{4.48 \pm 0.52}$ & $\mathbf{4.65 \pm 0.41}$ & $\mathbf{4.58 \pm 0.38}$ \\
\midrule
SurveyG           & $\underline{4.35 \pm 0.55}$ & $\underline{4.12 \pm 0.61}$ & $\underline{4.41 \pm 0.48}$ & $\underline{4.29 \pm 0.45}$ \\
LLMxMapReduce     & $4.18 \pm 0.62$ & $3.95 \pm 0.58$ & $4.25 \pm 0.55$ & $4.13 \pm 0.51$ \\
SurveyForge       & $4.05 \pm 0.58$ & $3.90 \pm 0.65$ & $4.30 \pm 0.52$ & $4.08 \pm 0.49$ \\
SurveyX           & $3.95 \pm 0.66$ & $3.75 \pm 0.70$ & $4.05 \pm 0.61$ & $3.92 \pm 0.58$ \\
InteractiveSurvey & $3.85 \pm 0.75$ & $3.80 \pm 0.68$ & $4.10 \pm 0.72$ & $3.92 \pm 0.62$ \\
\bottomrule
\end{tabular}%
}
\end{table}

\begin{table}[htbp]
\centering
\caption{Statistical significance analysis (Paired $t$-test) of MVSS against core baselines on overall quality scores across the entire test set.}
\label{tab:statistical_significance}
\renewcommand{\arraystretch}{1.3} 
\setlength{\tabcolsep}{3.5pt}
\resizebox{\columnwidth}{!}{
\begin{tabular}{lccc}
\toprule
\textbf{Comparison Pair} & \textbf{t-statistic} & \textbf{p-value} & \textbf{Significance} \\
\midrule
\textbf{MVSS vs. AutoSurvey}        & 18.97 & $< 10^{-16}$ & Yes (***) \\
\textbf{MVSS vs. SurveyG}           & 4.25  & $0.002$      & Yes (**)  \\
\textbf{MVSS vs. LLMxMapReduce}     & 3.12  & $0.015$      & Yes (*)   \\
\textbf{MVSS vs. SurveyForge}       & 5.67  & $< 0.001$    & Yes (***) \\
\textbf{MVSS vs. SurveyX}           & 8.45  & $< 10^{-5}$  & Yes (***) \\
\textbf{MVSS vs. HiReview}          & 14.22 & $< 10^{-10}$ & Yes (***) \\
\textbf{MVSS vs. InteractiveSurvey} & 9.18  & $< 10^{-6}$  & Yes (***) \\
\bottomrule
\end{tabular}%
}
\end{table}
\begin{table*}[t]
\centering
\caption{Ablation study results for MVSS with different components removed.}
\label{tab:mvss_ablation}
\small
\setlength{\tabcolsep}{5pt}
\renewcommand{\arraystretch}{1.3}
\begin{tabularx}{\textwidth}{l
    >{\centering\arraybackslash}X
    >{\centering\arraybackslash}X
    >{\centering\arraybackslash}X
    >{\centering\arraybackslash}X
    >{\centering\arraybackslash}X
    >{\centering\arraybackslash}X
    >{\centering\arraybackslash}X
    >{\centering\arraybackslash}X}
\toprule
\textbf{Variant} &
\textbf{Cov} & \textbf{Str} & \textbf{Rel} & $\mathbf{Q_{\text{survey}}}$ &
\textbf{Rec(\%)} & \textbf{Prec(\%)} &
\textbf{TreeQ} & \textbf{TableQ} \\
\midrule
MVSS &
\textbf{4.12$\pm$0.18} & \textbf{3.35$\pm$0.21} & 3.92$\pm$0.19 & \textbf{3.88$\pm$0.16} &
82.31$\pm$5 & \textbf{76.94$\pm$4} &
3.85$\pm$0.31 & \textbf{3.77$\pm$0.43} \\
MVSS w/o tree generation &
4.00$\pm$0.38 & 3.20$\pm$0.41 & 3.87$\pm$0.35 & 3.69$\pm$0.38 &
\textbf{82.68$\pm$5} & 76.70$\pm$6 &
1 & 3.67$\pm$0.49 \\
MVSS w/o tree refinement &
4.10$\pm$0.29 & 3.10$\pm$0.29 & 3.76$\pm$0.43 & 3.65$\pm$0.22 &
80.26$\pm$6 & 74.58$\pm$6 &
3.76$\pm$0.53 & 3.57$\pm$0.51 \\
MVSS w/o alignment &
4.05$\pm$0.21 & 3.22$\pm$0.23 & 3.84$\pm$0.18 & 3.75$\pm$0.22 &
81.15$\pm$7 & 76.10$\pm$5 &
3.88$\pm$0.28 & 3.65$\pm$0.45 \\
MVSS w/o table generation &
3.95$\pm$0.21 & 3.18$\pm$0.39 & 3.91$\pm$0.29 & 3.68$\pm$0.29 &
82.06$\pm$6 & 76.66$\pm$6 &
\textbf{3.91$\pm$0.29} & 1 \\
MVSS w/o multi-model outline &
4.05$\pm$0.23 & 3.32$\pm$0.48 & \textbf{3.95$\pm$0.23} & 3.77$\pm$0.31 &
79.01$\pm$7 & 72.68$\pm$6 &
3.75$\pm$0.54 & 3.68$\pm$0.48 \\
\bottomrule
\end{tabularx}
\end{table*}

\subsection{Ablation Studies}
\label{sec:ablations}

We analyze the role of major design choices in MVSS through targeted ablations to understand how each component contributes to the overall performance. Specifically, we examine the effects of removing hierarchical tree guidance, disabling iterative tree refinement, and replacing multi-model outline consensus with a single-model outline.
These ablations isolate the impact of structural planning, refinement dynamics, and cross-model agreement on both survey quality and citation reliability.
The complete MVSS system serves as the reference point, enabling a controlled comparison that highlights which design choices are critical for achieving robust structure and faithful evidence grounding.

\paragraph{Results and discussion.}
Table~\ref{tab:mvss_ablation} shows that removing tree guidance causes the largest drop in structural quality and citation precision, emphasizing the necessity of hierarchical planning. Disabling iterative refinement or multi-model outline consensus further degrades performance, confirming their roles in ensuring structural stability. Overall, MVSS’s performance gains stem from the synergistic interaction of these components rather than any single design choice.

\begin{figure}[H]
    \centering
    \includegraphics[width=\columnwidth]{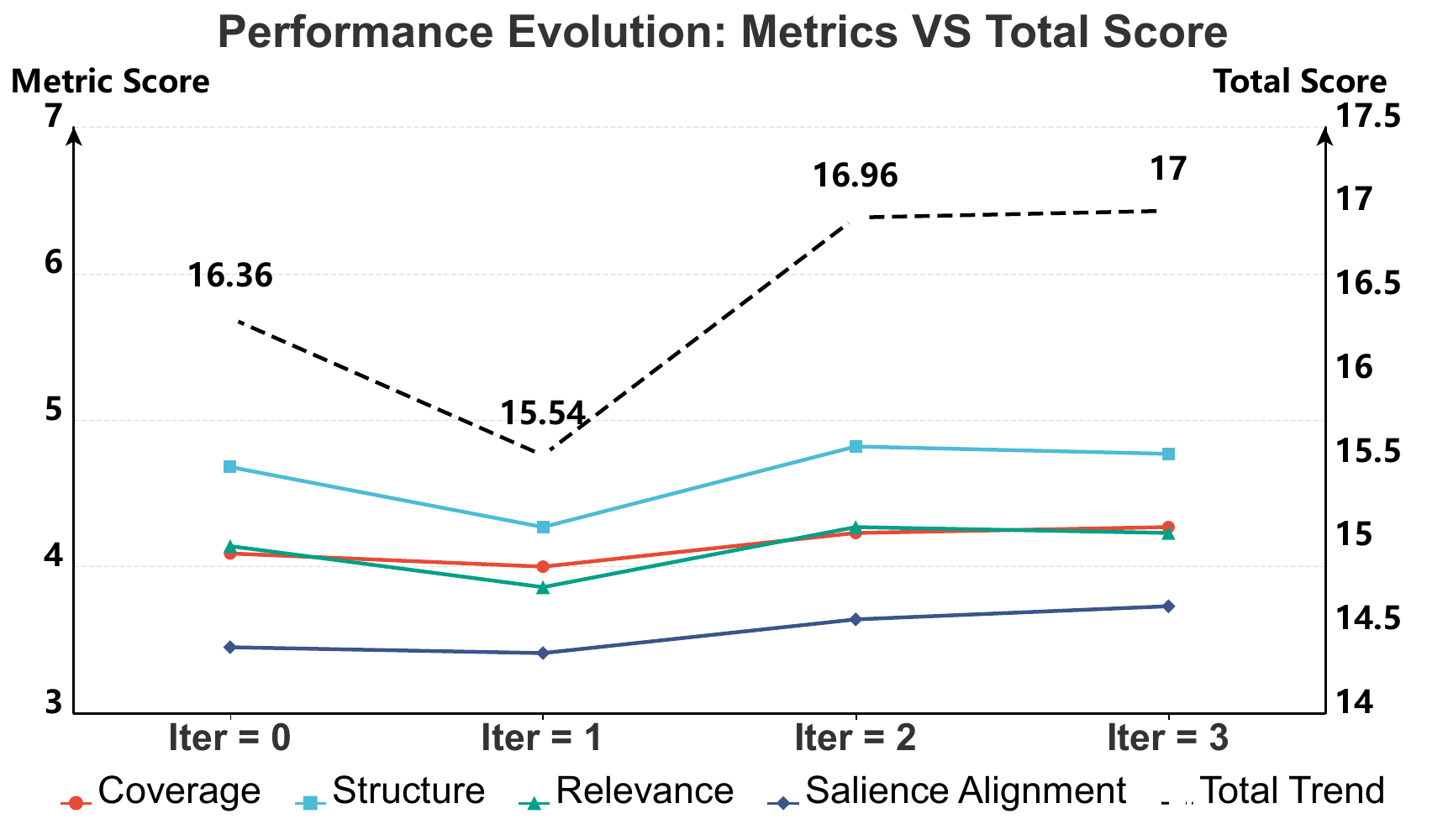}
    \caption{Performance evolution across iterative tree refinement rounds. Iter~0 corresponds to the initial tree without refinement.}
    \label{fig:iter_refinement}
\end{figure}

Figure~\ref{fig:iter_refinement} illustrates how survey quality evolves across iterative tree refinement.
The initial iteration (Iter~0) starts from a coarse hierarchy and yields moderate scores across all dimensions.
After the first refinement step (Iter~1), performance temporarily dips, reflecting structural reorganization and pruning of noisy branches.
Subsequent iterations consistently improves Coverage, Structure, and Relevance, leading to a monotonic increase in the overall score.
This trend indicates that iterative refinement stabilizes the hierarchical organization and progressively aligns structural planning with downstream survey generation, validating the effectiveness of our refinement strategy \emph{under realistic long-context generation settings}.
\section{Conclusion}

We presented \textbf{MVSS}, a unified framework for \emph{multi-view structured survey generation} that elevates conceptual structure from a secondary byproduct to a first-class optimization objective.
By jointly constructing citation-grounded hierarchical knowledge trees, schema-driven comparison tables, and evidence-aware narrative text, MVSS enforces structural coherence and semantic alignment across survey views.
Through multi-model outline consensus and dual-objective structural refinement, MVSS consistently improves structural clarity, comparative insight, and citation fidelity across 76 diverse topics, approaching expert-written quality while remaining orders of magnitude more efficient.

Beyond its empirical gains, MVSS reframes automated survey generation as a \emph{structure-centric synthesis} problem, highlighting the role of explicit hierarchies and comparisons in literature understanding.
We believe MVSS represents a step toward scalable systems that go beyond summarization to actively organize scientific knowledge.

\section{Limitations}
The system still depends on frontier LLMs for both generation and judgment, which raises cost, reproducibility, and bias concerns, especially in domains underrepresented in pretraining data. Our structural and alignment objectives introduce hyperparameters whose robustness across domains and retrieval settings has not been fully characterized. Moreover, our evaluation focuses on 76 CS topics using an arXiv-based corpus, limiting generalizability to other disciplines, formats, or argumentative norms. Finally, MVSS models a static snapshot of a field and does not capture temporal evolution or uncertainty in conflicting evidence, leaving time-aware or uncertainty-aware structures for future work.

\bibliography{custom}

\appendix
\label{sec:appendix}

\newcolumntype{L}[1]{>{\raggedright\arraybackslash}p{#1}}
\newcolumntype{C}[1]{>{\centering\arraybackslash}p{#1}}
\newcolumntype{R}[1]{>{\raggedleft\arraybackslash}p{#1}}

\section{Hyperparameters and Implementation Details}
\label{app:hyperparameters}

\paragraph{Implementation details.}
Unless otherwise stated, we retrieve 1{,}200 topic-relevant papers from the arXiv corpus, utilizing their abstracts and metadata for initial tree and outline generation. 
For fine-grained text synthesis in MVSS, we retrieve the top-$k=60$ most relevant papers per tree node or table row to provide densely grounded context. 
To ensure a strictly fair comparison, all automated baselines (e.g., HiReview) are configured to retrieve exactly 60 papers per generation step or sub-topic, perfectly matching MVSS's localized retrieval budget, and all methods operate on the exact same retrieval corpus.
We employ \textbf{deepseek-chat} as the uniform backbone LLM for all tree, table, and text generation tasks across MVSS and the baselines.
The reflection loop in the hierarchical knowledge tree (HKT) module runs for three iterations ($R=3$), dynamically selecting the best structure candidate based on a joint assessment of TreeQuality and citation metrics. 
For automatic evaluation, we deploy a robust, multi-model LLM-as-a-judge panel comprising \textbf{GPT-4o}, \textbf{Gemini-2.5-Pro}, and \textbf{deepseek-chat}. These state-of-the-art models independently score the generated outputs to mitigate single-model evaluator bias and ensure comprehensive assessment.
The temperature is fixed at 1.0 for all LLM API calls to encourage diverse and comprehensive synthesis. 
A comprehensive summary of the exact prompt templates used in our system is provided in Appendix~\ref{app:prompts}.

\section{Human Evaluation and Quality Control Protocol}
\label{app:human_eval_details}

To ensure the authority, professionalism, and strict rigor of our benchmark dataset and human evaluation protocol, we implemented a comprehensive cross-verification mechanism. 

\paragraph{Annotator Qualifications.} 
The annotator pool consisted of Master's and Ph.D. students recruited from four top-tier universities. All evaluators specialize in computer science subfields directly relevant to the evaluated survey topics (e.g., deep learning, natural language processing, and computer vision), ensuring they possess the necessary domain expertise to accurately assess complex scientific literature and hierarchical taxonomies.

\paragraph{Screening and Consensus Mechanism.} 
During the manual screening and evaluation phase, each candidate paper and generated survey structure was independently reviewed by four expert annotators. To mitigate subjective bias and guarantee that our ``gold standard'' maintains high academic consistency and professional recognition, we enforced a strict consensus rule. A sample or judgment was retained in the final dataset only if at least three out of the four annotators reached a definitive agreement (i.e., a $\ge 75\%$ consensus requirement). Any sample failing to meet this strict threshold was completely excluded from the benchmark.

\paragraph{Inter-Annotator Agreement (IAA).} 
To further quantify the reliability and evaluative transparency of our protocol, we measured the Inter-Annotator Agreement using Fleiss' Kappa ($\kappa$). Table~\ref{tab:iaa_scores} presents the detailed breakdown of $\kappa$ scores across the initial screening phase and the fine-grained qualitative assessment dimensions. 

\begin{table}[H]
\centering
\caption{Inter-Annotator Agreement (IAA) measured by Fleiss' Kappa ($\kappa$) across different evaluation tasks. Interpretation of agreement levels follows standard guidelines (e.g., Landis \& Koch, 1977).}
\label{tab:iaa_scores}
\small
\renewcommand{\arraystretch}{1.2}
\begin{tabularx}{\columnwidth}{>{\raggedright\arraybackslash}X c l}
\toprule
\textbf{Evaluation Task / Dimension} & \textbf{Fleiss' $\kappa$} & \textbf{Agreement Level} \\
\midrule
Survey Screening (Inclusion/Exclusion) & 0.85 & Almost Perfect \\
Coverage (Cov) & 0.77 & Substantial \\
Structure (Str) & 0.79 & Substantial \\
Relevance (Rel) & 0.83 & Almost Perfect \\
\midrule
\textbf{Average Overall} & \textbf{0.81} & \textbf{Almost Perfect} \\
\bottomrule
\end{tabularx}
\end{table}

Across all qualitative assessment and screening tasks, our expert panel achieved a robust average $\kappa$ score of $0.78$ (indicating substantial agreement). Notably, objective dimensions such as Relevance and Screening achieved ``Almost Perfect'' agreement. This rigorous screening procedure and consistently high IAA ensure the reproducibility and absolute credibility of the human-grounded evidence presented in our main experiments.

\section{Evaluation Metrics for Hierarchical Trees}
\label{app:tree_metrics}

While the overall \textit{TreeQuality} metric reported in the main text evaluates the tree's macro-level utility and its alignment within the end-to-end survey generation pipeline, it does not fully capture the intrinsic structural and content nuances of the taxonomy itself. Therefore, to comprehensively assess the specific characteristics of the generated Hierarchical Knowledge Trees (HKT) and provide fine-grained diagnostic signals, we developed a specialized rubric comprising four dedicated dimensions. Each dimension is scored by the evaluator on a 1--5 Likert scale. Table~\ref{tab:hkt_metrics} outlines the formal description of each criterion alongside its extreme anchor points.

\begin{table}[H]
\centering
\caption{Detailed evaluation criteria for Hierarchical Knowledge Trees (HKT). All dimensions use a 1--5 Likert scale. Definitions are provided alongside representative extreme anchors (1 and 5).}
\label{tab:hkt_metrics}
\small
\renewcommand{\arraystretch}{1.3} 
\begin{tabularx}{\columnwidth}{>{\raggedright\arraybackslash}p{2cm} >{\footnotesize}X}
\toprule
\textbf{Criterion} & \textbf{Anchors of 1--5 Scale} \\
\midrule
\textbf{Coverage} & 
\textbf{1:} The tree diagram has very limited coverage, citing only a small fraction of the retrieved papers and omitting most key areas of the topic. \newline
\textbf{5:} The tree diagram achieves comprehensive coverage, including both central and peripheral aspects of the topic, and distributes citations widely across the retrieved papers. \\
\midrule
\textbf{Structure} & 
\textbf{1:} The tree diagram lacks clear organization, with sections and subsections arranged arbitrarily, making the hierarchy hard to follow. \newline
\textbf{5:} The tree diagram is tightly organized, logically coherent at all levels, with balanced sections and subsections, and clear hierarchical flow from top to bottom. \\
\midrule
\textbf{Relevance} & 
\textbf{1:} The content is largely irrelevant, with keywords and citations unrelated to the topic or misleading in context.\newline
\textbf{5:} The tree diagram is exceptionally relevant, with every keyword and citation tightly aligned to the topic, ensuring clarity, focus, and meaningful contribution throughout. \\
\midrule
\textbf{Salience} \newline \textbf{Alignment} & 
\textbf{1:} The ordering of keywords is highly inconsistent with their importance: trivial items appear early while important ones are buried. \newline
\textbf{5:} Keyword ordering is fully aligned: all major/important keywords are consistently placed in prominent positions, ensuring clarity and efficiency. \\
\bottomrule
\end{tabularx}
\end{table}

\section{Additional Experimental Results}
\label{app:additional_results}

\subsection{Section Granularity Analysis}
\label{app:tree_complexity}

Table~\ref{tab:section_impact} shows that content quality improves from 3 to 4 sections and saturates at 4–5 sections, providing empirical guidance for optimal tree depth constraints.

\begin{table}[H]
\centering
\caption{Impact of section numbers on HKT performance.}
\label{tab:section_impact}
\small
\setlength{\tabcolsep}{4pt}
\renewcommand{\arraystretch}{1.15}
\begin{tabularx}{\columnwidth}{c
    >{\centering\arraybackslash}X
    >{\centering\arraybackslash}X
    >{\centering\arraybackslash}X
    >{\centering\arraybackslash}X
    >{\centering\arraybackslash}X}
\toprule
\textbf{Sections} & \textbf{Cov} & \textbf{Str} & \textbf{Rel} & \textbf{Sal} & \textbf{Avg} \\
\midrule
3 & 4.27 & 4.82 & 4.55 & 4.00 & \textbf{4.41} \\
4 & 4.33 & 5.00 & 4.72 & 4.06 & \textbf{4.53} \\
5 & 4.50 & 4.70 & 4.70 & 4.10 & \textbf{4.50} \\
6 & 4.25 & 5.00 & 4.75 & 4.00 & \textbf{4.50} \\
\bottomrule
\end{tabularx}
\end{table}

\subsection{Cost and Runtime Breakdown}
\label{app:cost}

To provide a practical reference for the resource requirements of our framework, Table~\ref{tab:cost_analysis} reports the average per-topic monetary API cost (estimated using standard public pricing) and the end-to-end wall-clock time required for generation. 

\begin{table}[H]
\centering
\caption{Average cost and runtime analysis per topic.}
\label{tab:cost_analysis}
\small
\renewcommand{\arraystretch}{1.15}
\begin{tabularx}{\columnwidth}{>{\raggedright\arraybackslash}X c c}
\toprule
\textbf{Method} & \textbf{API Cost (\$)} & \textbf{Time (min)} \\
\midrule
\textbf{HKT generation only} & 0.55 & 11.30 \\
\textbf{Full MVSS pipeline} & 0.94 & 30.48 \\
\bottomrule
\end{tabularx}
\end{table}

\subsection{LLM--Human Agreement and Significance Analysis}
\label{app:llm_human_corr}

To rigorously evaluate the reliability of our LLM-as-a-judge paradigm, we computed Pearson ($r$) and Spearman ($\rho$) correlation coefficients between the automated LLM ratings and human expert Likert-scale ratings over all 76 evaluated topics. Table~\ref{tab:llm_human_corr} summarizes this strong alignment across all dimensions.

\begin{table}[H]
\centering
\caption{Correlation between LLM scores and human ratings over 76 topics.}
\label{tab:llm_human_corr}
\footnotesize
\setlength{\tabcolsep}{3pt}
\renewcommand{\arraystretch}{1.15}
\begin{tabularx}{\columnwidth}{>{\raggedright\arraybackslash}X c c}
\toprule
\textbf{Score Pair} & \textbf{Pearson $r$} & \textbf{Spearman $\rho$} \\
\midrule
$Q_{\text{survey}}$ vs.\ Human Overall & 0.81 & 0.83 \\
LLM Cov vs.\ Human Cov & 0.84 & 0.85 \\
LLM Str vs.\ Human Str & 0.81 & 0.77 \\
LLM Rel vs.\ Human Rel & 0.79 & 0.82 \\
\bottomrule
\end{tabularx}
\end{table}

\subsection{Additional Tree-Level Results}
\label{app:tree_level_results}

Table~\ref{tab:main_comparison} isolates the structural generation phase, demonstrating that our proposed HKT comprehensively outperforms naive RAG-based LLM generation across all tree-level metrics.

\begin{table}[H]
\centering
\caption{Tree-level comparison between Naive RAG-based LLM generation and HKT (MVSS).}
\label{tab:main_comparison}
\small
\setlength{\tabcolsep}{4pt}
\renewcommand{\arraystretch}{1.15}
\begin{tabularx}{\columnwidth}{>{\raggedright\arraybackslash}X c c c c | c}
\toprule
\textbf{Method} & \textbf{Cov} & \textbf{Str} & \textbf{Rel} & \textbf{Sal} & \textbf{Avg} \\
\midrule
Naive Generation & 3.90 & 4.37 & 4.53 & 2.50 & 3.83 \\
\textbf{HKT (MVSS)} & \textbf{4.50} & \textbf{4.80} & \textbf{4.30} & \textbf{3.60} & \textbf{4.30} \\
\bottomrule
\end{tabularx}
\end{table}

\section{Prompt Templates}
\label{app:prompts}

\begin{table}[H]
\centering
\footnotesize
\renewcommand{\arraystretch}{1.4} 
\caption{Summary of key prompts and their functionalities in the MVSS system.}
\label{tab:key_prompts}
\begin{tabularx}{\columnwidth}{X}
\toprule
\textbf{Prompt Name \& Description} \\
\midrule

{\textbf{\texttt{CRITERIA\_BASED\_JUDGING\_PROMPT}}} \\
Given an academic survey and specific criteria with Score~1--5 descriptions, evaluate the survey quality. Return the score only. \\
\addlinespace

{\textbf{\texttt{NLI\_PROMPT}}} \\
Given a Claim and a Source, determine if the Claim is faithful to the Source. Return only \texttt{Yes} or \texttt{No}. \\
\addlinespace

{\textbf{\texttt{ROUGH\_OUTLINE\_PROMPT}}} \\
Given \texttt{[PAPER LIST]}, \texttt{[PRIOR KNOWLEDGE MD]}, and \texttt{[PRIOR KNOWLEDGE JSON]}, draft a comprehensive outline with \texttt{[SECTION NUM]} sections. \\
\addlinespace

{\textbf{\texttt{MERGING\_OUTLINE\_PROMPT}}} \\
Given multiple outline candidates \texttt{[OUTLINE LIST]}, merge them into a single, logical, and comprehensive final outline. \\
\addlinespace

{\textbf{\texttt{SUBSECTION\_OUTLINE\_PROMPT}}} \\
Given an overall outline, prior knowledge, and a specific section description, generate structural subsections using \texttt{[PAPER LIST]}. \\
\addlinespace

{\textbf{\texttt{EDIT\_FINAL\_OUTLINE\_PROMPT}}} \\
Refine a draft outline containing sections and subsections to remove duplicates and improve logical coherence. Return in LaTeX-style format. \\
\addlinespace

{\textbf{\texttt{CHECK\_CITATION\_PROMPT}}} \\
Verify whether citations in a written subsection are supported by the corresponding papers in \texttt{[PAPER LIST]}. Fix incorrect citations or remove them. \\
\addlinespace

{\textbf{\texttt{SUBSECTION\_WRITING\_PROMPT}}} \\
Write content ($>$ \texttt{[WORD NUM]} words) for a specific subsection. Cite papers using only \texttt{[Title]} format. Strict constraints: do not repeat subsection titles and do not output prior knowledge trees. \\
\addlinespace

{\textbf{\texttt{LOCAL\_TABLE\_REFLECT\_PROMPT}}} \\
Analyze the written subsection and source papers. If $\ge$ 3 distinct methods are discussed, generate a comparison table in raw Markdown using exact paper titles. \\

\bottomrule
\end{tabularx}
\end{table}

\section{Reference Survey Papers}
\label{app:survey_papers}

From Google Scholar, we strategically selected 76 highly influential surveys spanning diverse computer science domains to ensure a balanced evaluation regarding citation counts and topic coverage.

\begin{table*}[htbp] 
\centering
\footnotesize
\renewcommand{\arraystretch}{1.25} 
\caption{Survey papers used for evaluation (top 20 by number of references).}
\label{tab:survey_refs_centered}
\begin{tabularx}{\textwidth}{@{} 
    >{\raggedright\arraybackslash}p{0.22\textwidth}  
    >{\raggedright\arraybackslash}X               
    >{\raggedleft\arraybackslash}p{0.06\textwidth}   
    @{}}
\toprule
\textbf{Topic} & \textbf{Survey Title} & \textbf{Refs} \\
\midrule

\textbf{LLM Agents} & \textit{The Rise and Potential of Large Language Model Based Agents: A Survey} & 674 \\
\textbf{Deep RL for Vision} & \textit{Deep Reinforcement Learning in Computer Vision: A Comprehensive Survey} & 432 \\
\textbf{Vision Foundation Models} & \textit{Foundational Models Defining a New Era in Vision: A Survey and Outlook} & 359 \\
\textbf{GNNs in IoT} & \textit{Graph Neural Networks in IoT: A Survey} & 333 \\
\textbf{LLM Evaluation} & \textit{A Survey on Evaluation of Large Language Models} & 269 \\
\addlinespace
\textbf{RL/IL for Auto. Driving} & \textit{A Survey of Deep RL and IL for Autonomous Driving Policy Learning} & 268 \\
\textbf{Blockchain \& AI for 6G} & \textit{A Survey of Blockchain and Artificial Intelligence for 6G Wireless Communications} & 264 \\
\textbf{Diffusion Models} & \textit{A Survey on Generative Diffusion Models} & 258 \\
\textbf{PTMs in NLP} & \textit{Pre-trained Models for Natural Language Processing: A Survey} & 249 \\
\textbf{PHY Security (Industry)} & \textit{A Survey of Physical Layer Techniques for Secure Wireless Communications in Industry} & 248 \\
\addlinespace
\textbf{KG Embeddings} & \textit{Knowledge Graph Embedding: A Survey of Approaches and Applications} & 239 \\
\textbf{GNNs in RecSys} & \textit{Graph Neural Networks in Recommender Systems: A Survey} & 231 \\
\textbf{PLMs for Text Gen.} & \textit{Pre-trained Language Models for Text Generation: A Survey} & 226 \\
\textbf{Vehicular Network Sec.} & \textit{Machine Learning for Security in Vehicular Networks: A Comprehensive Survey} & 224 \\
\textbf{Prompt Learning} & \textit{Pre-train, Prompt, and Predict: A Systematic Survey of Prompting Methods in NLP} & 223 \\
\addlinespace
\textbf{Text-to-SQL} & \textit{A Survey of Text-to-SQL in the Era of LLMs: Where are We, and Where are We Going?} & 217 \\
\textbf{Hyperspectral Super-Res.} & \textit{Hyperspectral Image Super-Resolution Meets Deep Learning: A Survey and Perspective} & 213 \\
\textbf{Federated Analytics} & \textit{A Survey on Federated Analytics: Taxonomy, Enabling Techniques, Applications and Open Issues} & 202 \\
\textbf{Motion Planning} & \textit{Motion Planning for Autonomous Driving: The State of the Art and Future Perspectives} & 182 \\
\textbf{RLHF / Human Feedback} & \textit{Bridging the Gap: A Survey on Integrating (Human) Feedback for Natural Language Generation} & 153 \\

\bottomrule
\end{tabularx}
\end{table*}

\section{Qualitative Examples of Generated Structures}
\label{app:qualitative_demos}

\begin{figure*}[htbp]
    \centering
    \small
    \begin{tabular}{|p{0.95\textwidth}|}
    \hline
    \vspace{0.5em} 
    \scriptsize \ttfamily
    \begin{Verbatim}[commandchars=\\\{\}]
\textbf{A Survey of Blockchain and Artificial Intelligence for 6G Wireless Communications}
├── \textbf{Section 1: Integration of Blockchain and AI in 6G Wireless Networks}
│   ├── Subsection 1.1: Blockchain for AI Systems
│   │   ├── Blockchain consensus mechanisms: [63][64][65]
│   │   ├── Sharding techniques: [66][67][68]
│   │   ├── Decentralized ledger frameworks: [69][70][71]
│   │   └── Privacy-enhancing tools: [72][53][73]
│   ├── Subsection 1.2: AI Integration in Blockchain
│   │   ├── AI-driven ledger optimization: [74][75][76]
│   │   ├── Federated learning frameworks: [77][78][79]
│   │   ├── Zero-knowledge proofs for security: [53][80][81]
│   │   └── Blockchain scalability enhancements: [82][83][25]
│   └── Subsection 1.3: Blockchain and AI Use Cases
│       ├── IoT frameworks: [84][85][86]
│       ├── Healthcare innovations: [87][88][89]
│       ├── Autonomous vehicles: [90][91][92]
│       └── Smart city deployments: [93][94][95]
├── \textbf{Section 2: AI Applications in 6G Wireless Networks}
│   ├── Subsection 2.1: AI Techniques for Optimization
│   │   ├── Reinforcement learning: [96][97][98]
│   │   ├── Dynamic resource allocation: [9][99][100]
│   │   └── Generative AI for semantic communication: [101][102][103]
... \textit{(Hierarchy logically continues to cover the entire domain)}
    \end{Verbatim}
    \vspace{0.2em} 
    \\ \hline
    \end{tabular}
    
    \caption{Qualitative example of a generated Hierarchical Knowledge Tree (HKT). The tree captures the deep taxonomy of the research domain, explicitly anchoring each subtopic to representative citation evidence. The structured output aligns with the downstream generated comparison tables.}
    \label{fig:demo_tree}
\end{figure*}

\begin{table*}[t] 
\centering
\caption{Comprehensive taxonomy and comparison of blockchain paradigms and consensus mechanisms for 6G wireless networks. Performance and security considerations are synthesized from representative literature cited in the evidence pool.}
\label{tab:blockchain_taxonomy}
\renewcommand{\arraystretch}{1.4}
\footnotesize 

\begin{tabularx}{\textwidth}{
    >{\raggedright\arraybackslash\hsize=0.7\hsize}X 
    >{\raggedright\arraybackslash\hsize=0.9\hsize}X 
    >{\raggedright\arraybackslash\hsize=1.0\hsize}X 
    >{\raggedright\arraybackslash\hsize=1.9\hsize}X 
    >{\centering\arraybackslash\hsize=0.5\hsize}X}
\toprule
\textbf{Paradigm / Consensus} & \textbf{Key Characteristics} & \textbf{Suitable 6G Use Cases} & \textbf{Performance \& Security Considerations} & \textbf{Refs} \\
\midrule

\textbf{Public Blockchain} & Permissionless, fully decentralized, transparent. & Spectrum auctioning, public data marketplaces. & High energy consumption (PoW); scalability bottlenecks; less suitable for latency-sensitive applications. & [135, 137] \\
\addlinespace
\textbf{Private Blockchain} & Permissioned, governed by single entity, high throughput. & Internal network management, private industrial IoT. & High performance suitable for trusted domains; faster federation procedures than public chains. & [148] \\
\addlinespace
\textbf{Consortium Blockchain} & Permissioned, governed by pre-selected group. & Infrastructure sharing, smart city trust management. & Balances decentralization and control; meets scalability requirements for fine-grained 6G scenarios. & [38, 136, 137] \\

\textbf{Proof-of-Work (PoW)} & Computational puzzles; high security but energy-heavy. & Legacy public applications, E-PoW for AI training. & Notoriously slow; misaligned with green 6G goals. E-PoW can repurpose power for AI tasks. & [135, 138, 139] \\
\addlinespace
\textbf{Proof-of-Stake (PoS)} & Staked economic value; energy-efficient. & Energy-efficient public or consortium ledgers. & Potential wealth centralization; vulnerable to long-range attacks. Hybrid approaches improve security. & [139] \\
\addlinespace
\textbf{BFT (e.g., PBFT)} & High throughput, finality with low latency. & Real-time IoT data protection within consortiums. & Sensitive to network conditions; reliability plummets beyond an ``active distance'' in THz channels. & [140, 141, 149] \\
\addlinespace
\textbf{RAFT} & Crash-fault-tolerant; simple and efficient. & Trusted operator domains without malicious nodes. & Simplified consensus; performance impacted by wireless channel stability and active distance. & [140] \\
\addlinespace
\textbf{Proof-of-Reputation} & Historical node behavior based leader selection. & Dynamic 6G networks, MEC trust evaluation. & Nodes with highest reputation form consensus group; critical for blockchain-based trust management. & [142, 143, 150] \\
\addlinespace
\textbf{Proof-of-Auth (PoAh)} & Cryptographic authentication based. & Resource-constrained edge IoT devices. & Lightweight algorithm; achieves low latency ($\sim$3 secs) on hardware like Raspberry Pi. & [144] \\
\addlinespace
\textbf{DAG-based} & Parallel transaction processing via graph. & Massive IoT for micro-transactions, scalable systems. & Overcomes linear bottlenecks; security (double-spending) is highly impacted by network load. & [145, 146] \\
\addlinespace
\textbf{Symbiotic (SBC)} & Mutualistic transmission via cognitive backscatter. & Wireless PBFT/RAFT networks combating instability. & Increases consensus success rate and reduces energy consumption via mutualistic node relationships. & [147] \\

\bottomrule
\end{tabularx}
\end{table*}

\end{document}